# A Training-Free Framework for Open-Vocabulary Image Segmentation and Recognition with EfficientNet and CLIP

Ying Dai, *Member, IEEE*, Wei Yu Chen

*Abstract*— **This paper presents a novel training-free framework for open-vocabulary image segmentation and object recognition (OVSR), which leverages EfficientNetB0, a convolutional neural network (CNN), for unsupervised segmentation and CLIP, a vision–language model, for open-vocabulary object recognition. The proposed framework adopts a two-stage pipeline: unsupervised image segmentation followed by segment-level recognition via vision-language alignment. In the first stage, pixel-wise features extracted from EfficientNetB0 are decomposed using singular value decomposition (SVD) to obtain latent representations, which are then clustered using hierarchical clustering to segment semantically meaningful regions. The number of clusters is adaptively determined by the distribution of singular values. In the second stage, the segmented regions are localized and encoded into image embeddings using the Vision Transformer (ViT) backbone of CLIP. Text embeddings are precomputed using CLIP's text encoder from category-specific prompts, including a generic "something else" prompt to support open-set recognition. The image and text embeddings are concatenated and projected into a shared latent feature space via SVD to enhance cross-modal alignment. Recognition is performed by computing the softmax over the similarities between the projected image and text embeddings. The proposed method is evaluated on standard benchmarks, including COCO, ADE20K, and PASCAL VOC, achieving state-of-the-art performance in terms of Hungarian mIoU, precision, recall, and F1-score. These results demonstrate the effectiveness, flexibility, and generalizability of the proposed framework.**

*Index Terms*— **open-vocabulary, pixel-wise latent features, pixel-wise clustering, object localization, CLIP, object recognition**

## I. INTRODUCTION

The objective of open-vocabulary image segmentation and recognition is to enable models to segment and identify objects in images using arbitrary category names, including those not encountered during training. This task bridges vision and language by leveraging pretrained vision-language models, facilitating scalable and flexible semantic understanding without dependence on fixed label sets or task-specific annotations.

Image segmentation and recognition have two different categories: single-stage approaches and two-stage approaches. The former directly outputs each pixel-level mask and its label, and the latter decouples the problem into a class-agnostic segmentation task and a pixel-level mask classification task. However, an intuitive observation is that, when given an image, humans naturally group pixels into objects first and then assign semantic labels to them. For instance, a child can easily group the pixels of an object even without knowing its name [1]. Two-stage approaches explicitly model this object-centric grouping, enabling more human-aligned interpretation than single-stage methods [3]. As a result, they are more commonly employed in open-vocabulary object detection tasks [2]. Accordingly, this paper adopts a two-stage pipeline of image segmentation followed by segment-level recognition. This paper proposes a novel framework for unsupervised image segmentation and open-vocabulary object recognition by integrating EfficientNetB0, a convolutional neural network (CNN), with CLIP, a vision-language model.

The state-of-the-art (SOTA) research regarding two-stage image segmentation in the literature includes [3-7]. The authors in [3] introduced an image segmentation task which aimed to segment all visual entities (objects and stuffs) in an image without predicting their semantic labels. This paper proposed a CondInst[4]-like fully-convolutional architecture with two novel modules specifically designed to exploit the class-agnostic and non-overlapping requirements of entity segmentation. The authors in [5] proposed to decouple the zero-shot semantic segmentation (Z3S) into two subtasks and train a Z3S model in the two-stage approach. Authors in [6] introduced a segment anything model (SAM) which was designed and trained to be promptable. The authors in [7] proposed a mask classification model with predict a set of masks, each associated with a single global class label prediction. The two-stage method simplified the landscape of approaches to semantic segmentation tasks and showed good results. All these papers use the ResNet [12] backbones or vision transformer (ViT)-based Swin-Transformer [13] to obtain feature representations of image pixels for generating a set of masks. However, whether the ResNet or Swin-Transformer are the best backbones and whether there are other more effective backbones and methods to represent the pixels are not explored. Furthermore, the papers [3-6] used pixel-level manually-annotated masks to train the segmentation model with the mask loss. These

This paper is submitted to IEEE Trans. on Multimedia on xxx for review. This work was supported in part by The Japan Society for the Promotion of Science (JSPS) KAKENHI under Grant JP22K12095, and in part by the Japan Keirin Autorace Foundation (JKA) Subsidy Program for Keirin and Auto Racing. The corresponding author is Ying Dai.

Ying Dai, Iwate Prefectural University, Takizawa, Iwate, 0202-0693, Japan (e-mail: dai@iwate-pu.ac.jp).

Wei Yu Chen, Iwate Prefectural University, Takizawa, Iwate, 0202-0693, Japan (e-mail: s231x601@s.iwate-pu.ac.jp).

segmentation methods heavily depend on the human-labeled data which is both labor-intensive and costly and thus less scalable [8]. The paper [7] utilized the pre-trained transformer as a decoder to obtain the mask predictions. However, whether the transformer is optimal as a decoder is unclear.

Due to the limitations of pixel-level supervised semantic segmentation, several recent works have proposed promising approaches for unsupervised image segmentation. For instance, [10] employed self-supervised models such as DINO [9] to "discover" objects and train detection and segmentation models via a cut-and-learn pipeline. In [11], the semantically rich feature correlations produced by DINO were leveraged for segmentation using the k-nearest neighbors (KNN) algorithm. The authors of [8] clustered instance-level features from [10] using k-means to generate pseudo semantic labels, which were then used to train a segmentation model. Similarly, [12] extracted DINO features from foreground regions to train a semantic segmentation network with pseudo-labels, enabling segmentation in multi-object scenes. However, all these methods rely on DINO features obtained through self-supervised training, which limit their flexibility and scalability in open-vocabulary image segmentation scenarios.

In [13], the authors proposed DiffCut, a method that achieves zero-shot semantic segmentation by leveraging diffusion features and recursive normalized cuts. In [14], an approach was introduced that explores effective latent feature representations of image pixels using a convolutional neural network (CNN) to improve the unsupervised segmentation of ingredients in images. While these approaches can segment objects without requiring additional model training, they do not provide semantic labels for the segmented regions.

Recently, leveraging vision-language models (VLMs) such as CLIP to enable open-vocabulary image segmentation and recognition has become a prominent research topic. In [15], Authors designed a minimal set of changes to endow these VLMs with perceptual grouping, resulting in zero-shot segmentation without training on any segmentation data or performing task specific fine-tuning. Same to [13, 14], it does not provide semantic labels for the segmented regions.

In [16], a training approach leveraging a VLM was proposed to distill knowledge from a pretrained open-vocabulary image classification model into a two-stage object detector. In [17], authors proposed a simple yet effective framework to Distill the Knowledge from the VLM to a DETR-like detector, termed DK-DETR. In [18], authors proposed a hierarchical semantic distillation framework named HD-OVD to construct a comprehensive distillation process, which exploits generalizable knowledge from the CLIP model. However, all these approaches require training student models to enable object detection.

In [19], the authors present a method that classifies masked regions by fine-tuning CLIP combined with mask prompt tuning. Similar to [16-18], this method necessitates retraining CLIP. In [20], the pseudo-labels generated by the VLM are refined, and the detector weights are adaptively updated to suppress biased training boxes that are not well aligned with target objects. Similar to the aforementioned approaches, this method still requires a self-supervised training procedure.

In this paper, we present a novel two-stage training-free framework for open-vocabulary image segmentation and object recognition (OVSR). In the first stage, pixel-wise features extracted from EfficientNetB0 are decomposed via singular value decomposition (SVD) to obtain compact latent representations, which are subsequently clustered using hierarchical clustering to produce semantically meaningful segments. The number of clusters is adaptively determined by the distribution of singular values. In the second stage, the segmented regions are localized and encoded into image embeddings using the Vision Transformer (ViT) backbone of CLIP, while text embeddings are precomputed from category-specific prompts, including a generic "something else" prompt for open-set recognition. The image and text embeddings are concatenated and further projected into a shared latent space using SVD to enhance cross-modal alignment. Final recognition is performed by applying a softmax over the similarity scores between the projected image and text embeddings.

Our contributions are summarized as follows:
- Pixel-level feature representation: We construct pixel-level features by concatenating multi-scale feature maps extracted from different convolutional layers of EfficientNetB0 and decomposing them via SVD to obtain a compact latent space.
- Unsupervised segmentation: We segment images by performing hierarchical clustering on the pixel-wise latent features, where the number of clusters is automatically determined based on the singular value spectrum.
- Open-vocabulary recognition: We encode cropped regions and category vocabularies using CLIP, and project the resulting embeddings into a shared latent space for object recognition.
- Performance validation: We validate our approach on multiple benchmarks and achieve performance that surpasses SOTA methods.

## II. RELATED WORKS

### A. Unsupervised image segmentation

Unsupervised image segmentation seeks to partition an image into semantically coherent regions without relying on annotated labels, making it a challenging yet crucial problem in computer vision. Recent advances leverage deep representations from pretrained CNNs or vision transformers to capture high-level semantics, followed by clustering or self-supervised learning to produce pixel-level groupings. This paradigm enables scalable segmentation for open-world scenarios, particularly when labeled datasets are scarce or unavailable, and forms a foundation for many downstream vision tasks. In this paper, we extend the approach presented in [14], an unsupervised method for segmenting ingredients in food images, to segment generic objects in images.

*B. CLIP*

Contrastive Language-Image Pretraining (CLIP) [21] is a vision-language model that learns a joint embedding space for images and text through large-scale contrastive training on image-text pairs. By aligning visual features from a ViT [22] or ResNet [23] backbone with textual embeddings from a Transformer-based text encoder, CLIP enables zero-shot recognition through similarity comparison between image and text embeddings. For text encoding, the input sequence is bracketed with [SOS] and [EOS] tokens. The activation at the [EOS] token from the final transformer layer is used as the text representation, which is then layer-normalized and linearly projected into the multi-modal embedding space. Unlike conventional supervised models restricted to predefined categories, CLIP generalizes to open-vocabulary tasks by leveraging natural language prompts as flexible labels. This capability has made CLIP a powerful foundation model for image classification, segmentation, and cross-modal retrieval without task-specific training. In this paper, we leverage the encoders of CLIP to extract image and text embeddings.

*C. Benchmarks*

Microsoft COCO [25], ADE20K [26], and PASCAL VOC [27] are widely used benchmarks for evaluating image segmentation and object recognition tasks. COCO provides 5000 validation images with instance segmentation, object detection, and captioning annotations across 80 object categories, emphasizing objects in complex, real-world contexts. ADE20K provides 2000 validation images labeled with 150 categories, including objects, parts, and background regions, making it ideal for semantic segmentation. PASCAL VOC provides 2900 validation images with 20 object categories, offering pixel-level segmentation and detection annotations. In this paper, we leverage these three benchmarks to evaluate the performance of the proposed framework for open-vocabulary image segmentation and object recognition.

## III. OVERVIEW OF THE FRAMEWORK

The training-free framework addresses unsupervised image segmentation and open-vocabulary object recognition through a two-stage pipeline, shown as in Fig. 1. In the first stage, pixel-level features are extracted from EfficientNetB0 [24] and decomposed via singular value decomposition (SVD) to obtain compact latent representations. These representations are then clustered using hierarchical clustering to produce semantically meaningful segments, with the number of clusters adaptively determined based on the singular value spectrum. In the second stage, the segmented regions are encoded using the image encoder of CLIP, while category-specific text prompts are embedded using the text encoder. These cross-modal embeddings are aligned, and the cross-modal similarity is computed for open-vocabulary recognition. The following sections separately detail the approaches to unsupervised image segmentation and open-vocabulary object recognition.

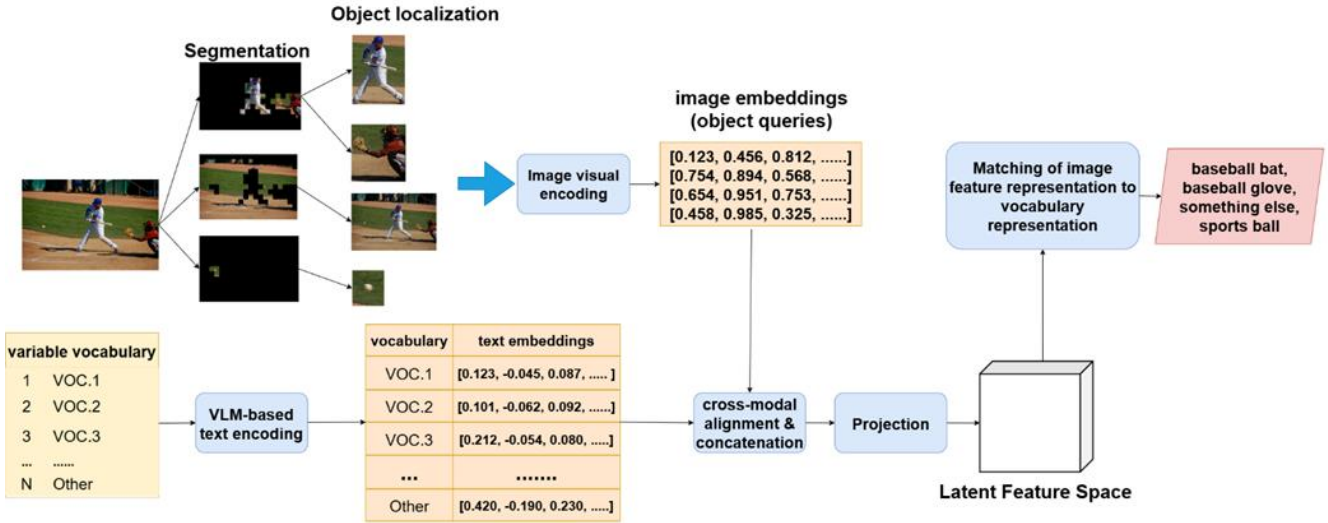

Fig. 1 Diagram of open-vocabulary image segmentation and recognition framework

## IV. UNSUPERVISED IMAGE SEGMENTATION

In [14], an unsupervised method for segmenting food ingredients was proposed. Building on the observations in that study, authors identify several key insights for image segmentation. First, conventional backbones such as ResNet and SAM [28] are suboptimal for generating discriminative feature representations across diverse images with respect to clustering quality, whereas EfficientNetB0 demonstrates superior performance in this regard. Second, constructing pixel-wise latent feature representations by concatenating multi-layer convolutional feature maps, followed by singular value decomposition (SVD), enhances inter-cluster separability and intra-cluster compactness for ingredient segmentation. Third, the number of clusters can be adaptively estimated from the singular value spectrum if the latent feature matrix is used to cluster image pixels, as the top singular values capture the dominant variance structure of the latent feature matrix.

Inspired by these insights, we propose a framework for unsupervised image segmentation. An overview is presented in Fig. 2.

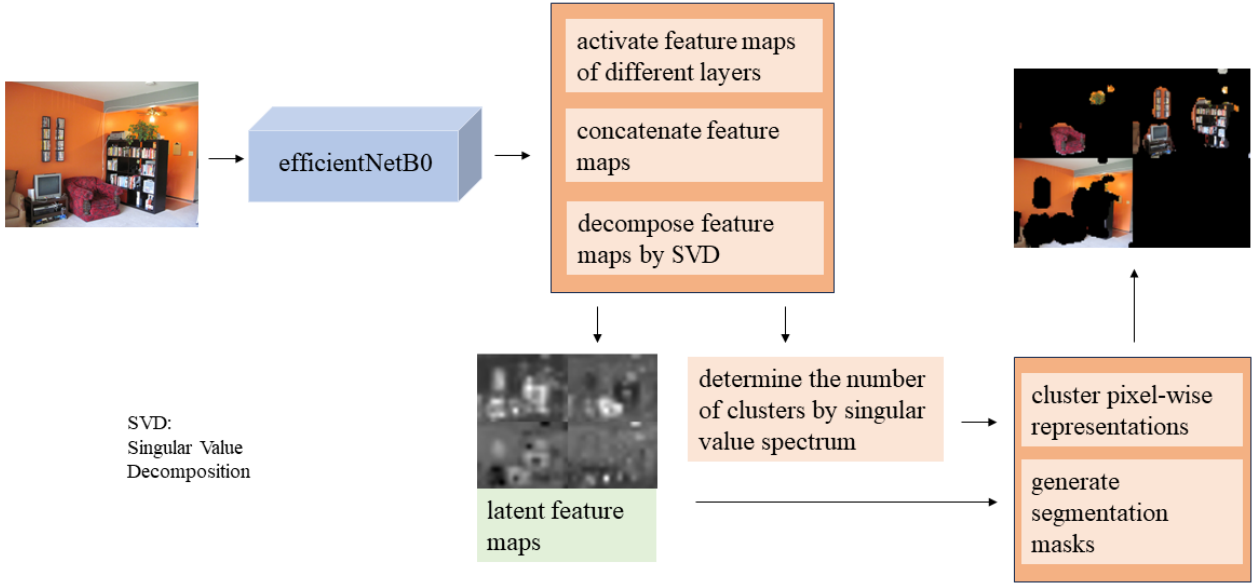

Fig.2 Overview of unsupervised image segmentation framework

According to this framework, images are fed into EfficientNetB0. EfficientNetB0 consists of a stem, 16 MBConv blocks (mobile inverted bottleneck blocks with SE modules), and a head, with Swish activations applied throughout. The stem outputs a tensor containing primitive features of colors; the last block outputs a tensor that is highly abstract, semantically rich, and spatially preserved; and the head outputs class-discriminative features representing entire objects or categories with global context. Moreover, Swish activations are implemented via an elementwise multiplication layer (MulLayer), which is defined as:

$Swish(A) = x * \sigma(A)$       (1)

Where: $A$ = input tensor, and $\sigma(A)$ = sigmoid function = $1/1+e^{-A}$.

Swish activations allow small negative values to pass, enabling richer feature representations. Its non-monotonicity facilitates better separation of fine-grained patterns. Together, these properties help the system extract more informative and discriminative features, particularly in challenging regions containing small objects.

Based on the above explanation, the Swish activations from the stem and the last block are used as the feature maps of the input images, because we assume that the features of colors and semantically rich features are play crucial roles in image segmentation. For capturing the intrinsic structure of the data, characterizing the variance distribution, and facilitating effective dimensionality reduction, these feature maps are then concatenated, flattened, normalized and decomposed using singular value decomposition (SVD). SVD decomposes a matrix into orthogonal directions with associated strengths (singular values). For any real matrix $A$ of size $m \times n$, the SVD is:

$A = U\Sigma V^{\mathrm{T}}$       (2)

Where: $U$: an $m \times m$ orthogonal matrix (left singular vectors); $V$: an $n \times n$ orthogonal matrix (right singular vectors); $\Sigma$: an $m \times n$ diagonal matrix whose diagonal entries are singular values ($\sigma_1 > \sigma_2 > \cdots \sigma_r > 0, r\ is\ the\ rank\ of\ \Sigma$). So, if the first $k$ singular vectors and top-$k$ columns of $V$ denoted as $V_k$ are kept, multiplying original data $A$ (or sample $x$) by $V_k$ can get their projection values in the new coordinate system, which are equivalent to $U\Sigma$, i.e. the data expressed along the SVD principal directions. The projection of samples onto the new basis vectors in $V$ forms latent feature representation space of the images. These representations are then reshaped to get the latent feature maps. Fig. 3 is an example showing top 12 latent feature maps of an image after SVD. Furthermore, plots of top 20 singular values of this image and their second-order differences are also exhibited in Fig. 3.

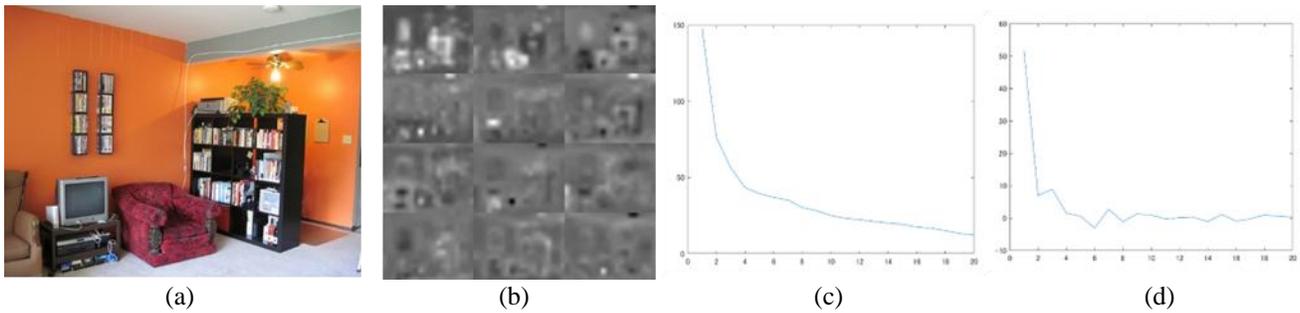

(a): Original image; (b): Top 12 latent feature maps; (c): Plot of singular values; (d): Plot of second-order differences

Fig.3 A example of latent feature maps

We observed that the first three feature maps predominantly attend to the main objects in the image, while the subsequent two highlight smaller objects; the remaining maps exhibit no salient attention patterns. Furthermore, we find from the plots that there are large changes in top 5 singular values. The position of the first slowdown point in the monotonic sequence of singular values is 6, where the second-order difference becomes negative, meaning the single value curve bends downward and incremental gains diminish. These observations imply that top latent feature maps of the images capture the intrinsic structure of the data, characterizing the variance distribution, which can facilitate effective dimensionality reduction. Accordingly, we determine to use top $k$ latent feature maps as pixel-wise representations, while the value of $k$ corresponds to the position of the first slowdown point in the singular value spectrum, minus one. Furthermore, we assume that setting $k$ as the number of clusters is appropriate for the subsequent unsupervised image segmentation. For this image, $k=5$.

According to [14], agglomerative hierarchical clustering is more robust to the lack of intra-cluster compactness caused by variations in colors and textures within the same object category, although it remains sensitive to noise and outliers. In this paper, agglomerative hierarchical clustering is used to segment images because it is more crucial to segment objects with varying colors and textures than to focus on noisy regions, which can be removed through post-processing. After clustering the pixel-wise representations, the resulting $k$ clusters are reshaped to form segmentation masks. Fig.4 is an example of image segmentation with the same representations and the different number of clusters. The result shown in (c) is the one with the number of clusters gotten by the proposed method.

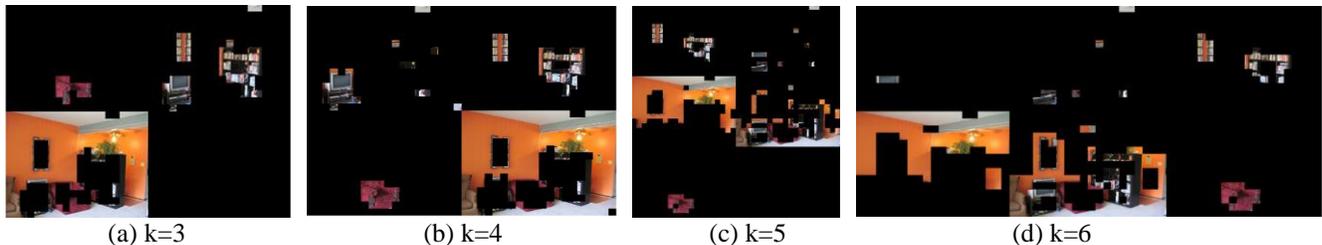

(a) k=3          (b) k=4          (c) k=5          (d) k=6

Fig. 4 A example of image segmentation with the different number of clusters

From these results, case (b) and case (c) yield relatively accurate object segmentation, whereas case (a) suffers from under-segmentation and case (d) from over-segmentation. Accordingly, these results validate the above assumption that the position of singular value spectrum can be used as the number of clusters to complete the unsupervised image segmentation dynamically.

As the whole, the process of unsupervised image segmentation is summarized as the following.

- Step 1. Deriving feature maps from Swish activations of stem and block 16 of EfficientNetB0;
- Step 2: Resizing and Concatenating feature maps to improve element-wise clustering performance for segmentation
- Step 3: flattening these feature maps to get pixel-wise representations;
- Step 4: Normalizing and singular value decomposing representations to get the latent feature representations with the number of clusters
- Step 5: Clustering the representations into clusters with the above number;
- Step 6: Reshaping the clusters and generating segmentation masks.
- Step 7: Obtaining segment images by element-wise multiplication of the masks and the original image.

Fig. 5 illustrates the benefits of resizing the feature maps in Step 2 for improved segmentation.

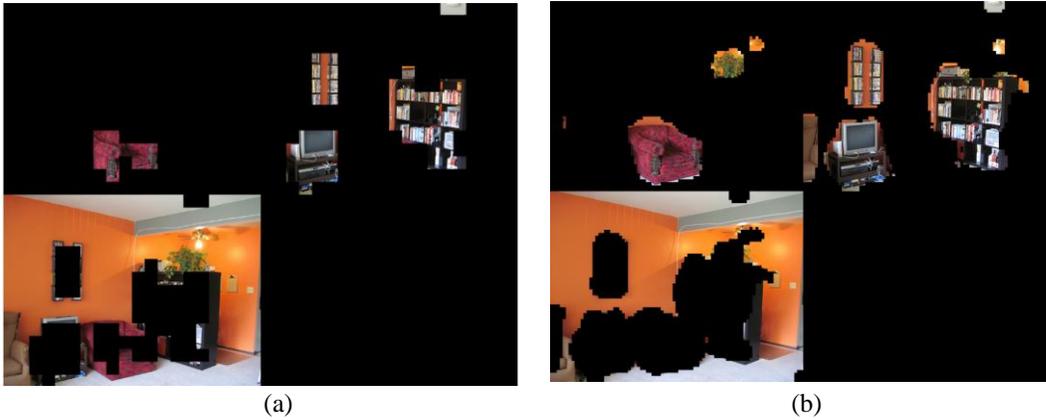

<div align="center">(a)               (b)</div>

(a): Segmentation results without resizing feature maps; (b): Segmentation results with resizing feature maps to 64×86

Fig. 5 A example of image segmentation with and without resizing feature maps

The original image has a resolution of 480×640, and the corresponding feature maps at block 16 of EfficientNetB0 exhibit a spatial size of 15×20. At this scale, the generated segments obtained using masks derived from element-wise clustering of the concatenated feature maps are illustrated in Fig. 5(a). When the feature maps are resized to 64×86, the resulting segments, shown in Fig. 5(b), demonstrate noticeably improved segmentation quality, as evidenced by clearer object boundaries and reduced under-segmentation artifacts. The segmentation in Fig. 5(b) outperforms that in Fig. 5(a), indicating that increasing the feature resolution to 64×86 enhances local detail representation and facilitates more accurate edge delineation. These results suggest that appropriately chosen feature-map resolutions preserve finer spatial cues, thereby contributing to more precise and visually consistent segmentation performance, albeit at the cost of increased computational complexity. The processing time in (a) is 2.69 seconds, whereas that in (b) is 3.94 seconds.

<div align="center">V. OPEN-VOCABULARY OBJECT RECOGNITION</div>

### A. Object Localization

In the open-vocabulary object recognition framework, all of the above semantic segments are utilized for object localization. As shown in Fig. 6, each segmentation map is resized to match the original image to ensure precise alignment between object regions and the corresponding image pixels. For reducing noise and enhance the quality of localized regions, the segmentation map is converted to grayscale and binarized, followed by morphological operations (including opening and dilation) to remove small artifacts and slightly expand object boundaries.

Connected-component analysis is then applied to identify each object region within the binary mask. Regions with an area smaller than a predefined threshold are discarded to avoid detecting noise or irrelevant regions. For each valid region, the bounding box coordinates are calculated and used to crop the corresponding object from the original image. Then, the cropped object images are encoded for object recognition in the subsequent step.

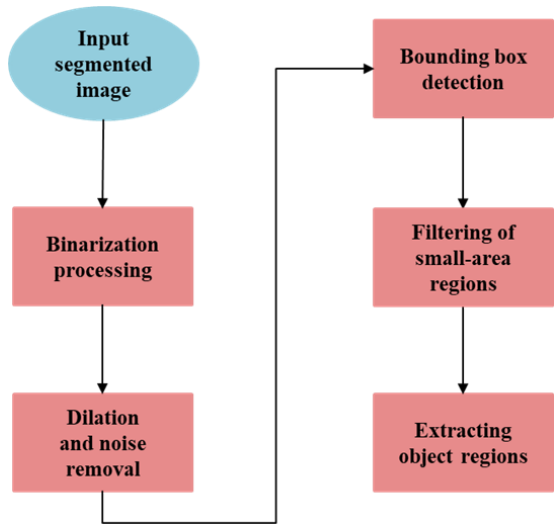

Fig. 6 Flowchart of object localization

*B. Embedding*

For text embeddings, we utilize the text encoder from the CLIP (Contrastive Language–Image Pretraining) model to convert category labels into semantic vectors. Since CLIP jointly represents images and text within a shared semantic space, it is highly suitable for open-vocabulary object recognition tasks.

To analyze the impact of prompt design on semantic representation and recognition performance, three types of textual prompts are designed as follows:

- Phrase 1: "a photo of a {super category} such as {category}" — emphasizes the semantic relationship between super-category and fine-grained class to improve generalization.
- Phrase 2: "this is a {category} of a {super category}" — provides a balanced and descriptive sentence structure, allowing the model to capture both categorical and contextual meaning.
- Phrase 3: "a photo of {category}" — the simplest form, excluding the super-category, allowing the model to focus purely on class-level semantics.

These prompt templates enable systematic analysis of how linguistic structure influences CLIP's semantic understanding and classification precision, particularly regarding whether including a super-category aids or hinders fine-grained recognition.

For object image embeddings, the CLIP image encoder (ViT-B/32 backbone) is used to convert cropped object regions into vector representations. Each input image is preprocessed following CLIP's standard pipeline of resizing and normalization before being passed through the encoder to obtain its feature embedding. Similar to text embeddings, all image embeddings are L2-normalized to lie on a unit hypersphere, facilitating cross-modal similarity computation.

To align the two modalities, the image and text embeddings are concatenated into a unified multimodal representation. The fusion preserves both linguistic semantics and visual appearance features, improving generalization in open-vocabulary recognition. The combined embeddings are then used for similarity computation and classification.

*C. Singular Value Decomposition (SVD)*

To remove redundant information from the embedding space, SVD is applied. By retaining only the top $k$ singular values and their corresponding vectors, a low-rank approximation is obtained. This low-rank representation preserves the essential structure of the data while removing minor or noisy components, resulting in a more compact and discriminative embedding.

In this study, both text and image embeddings are represented as high-dimensional matrices after feature extraction. For clip, the dimension of the embedding is 512. These matrices are individually processed by SVD, and the top $k$ components (based on singular values) are selected as the final reduced representation. The dimensionality reduction parameter $k$ was determined based on the number of categories in the dataset if this is significantly less than the embedding, so as to allow the latent feature space to maintain a comparable scale to the semantic diversity of the evaluation set. Since both text and image embeddings undergo the same SVD transformation, their reduced representations remain aligned within the same latent semantic space, enhancing the stability of cross-modal matching.

*D. object-category matching*

After SVD, the image and text embeddings are projected into a shared semantic space for object-category matching and classification. In this process, each segmented object is assigned to its most probable category. Fig. 7 presents a flowchart of the overall procedure.

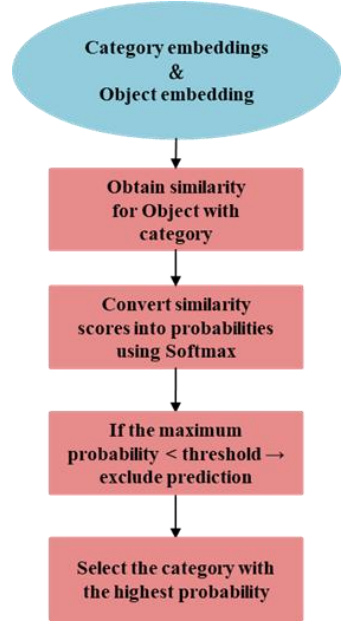

Fig. 7 Flowchart of object-category matching

First, cosine similarity is computed between each object embedding and all category embeddings to measure their semantic relatedness. The cosine similarity value ranges from −1 to 1, with higher values indicating stronger semantic similarity. This yields a similarity vector representing the relationship between each object and all candidate categories.

Next, the similarity scores are transformed into probabilities via the Softmax function as Fig. 8. Softmax performs exponential scaling and normalization over all class scores so that each output lies between 0 and 1, and all probabilities sum to 1. Through this transformation, raw similarity values become comparable probabilities. For example, given input scores [1.3,5.1,2.2,0.7,1.1], Softmax converts them to [0.02,0.90,0.05,0.01,0.02], indicating that the second class has the highest probability and is selected as the final prediction.

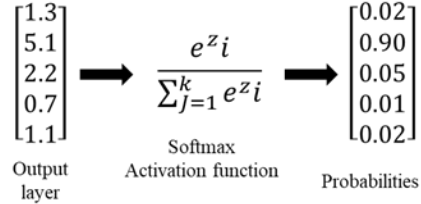

Fig. 8 Illustration of the Softmax function operation

To suppress false predictions, a fixed threshold $\theta$ is applied after Softmax. Predictions whose maximum probability falls below $\theta$ are discarded, ensuring reliable classification. This process, shown in the Fig. 7, effectively filters uncertain results and improves overall performance in open-vocabulary object recognition.

## VI. Experiments and analysis

We conduct evaluation studies using MATLAB 2024b on a Dell laptop equipped with a 12GB RTX 4090 Laptop GPU and a 64GB RAM.

### A. Evaluation metrics

We use Hungarian Intersection over Union (HIoU), accuracy, precision, recall, F1-score and AP to evaluate the performance of image segmentation and object recognition. Hungarian IoU evaluates class-agnostic segmentation by optimally matching predicted and ground-truth masks using the Hungarian algorithm. It builds an IoU matrix, finds one-to-one assignments maximizing total overlap, and computes mean IoU over matched pairs, ensuring fair, label-free comparison. the HIoU expression (for class-agnostic segmentation) is usually written as:

$$HIoU = \frac{1}{|\sigma^*|} \sum_{(i,j) \in \sigma^*} \frac{|P_i \cap G_j|}{|P_i \cup G_j|} \qquad (3)$$

Where, $P_i$ denotes predicted mask $i$, $G_j$ denotes ground-truth mask $j$, $\sigma^*$ denotes optimal one-to-one assignment from the Hungarian algorithm maximizing total IoU, and $|\sigma^*|$ equals to the number of $(i,j)$ pairs such that $P_i$ is matched to $G_j$.

Accuracy represents the overall proportion of correctly classified samples. Precision measures the proportion of correctly predicted samples among all positive predictions, while Recall measures the proportion of correctly retrieved samples among all actual positives. Since these two metrics often trade off against each other, their harmonic mean (F1-score) is adopted to evaluate overall balance. For AP, conventional object detection uses IoU-based region overlap; however, this study performs semantic matching rather than spatial localization. Therefore, AP is defined as the product of Precision and Recall, approximating the model's overall semantic recognition ability.

## B. Segmentation results

We conduct segmentation experiments using the proposed method on the validation datasets of the COCO-Obj-80, ADE20K, and PASCAL VOC benchmarks. Fig. 9 shows an example of segmentation for an image from COCO, with an *HIoU* of 44.9%.

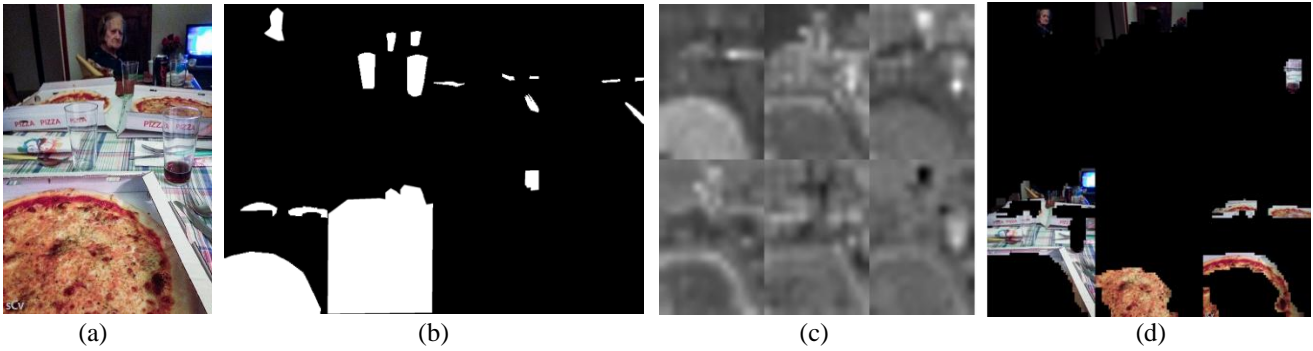

(a): Original image; (b): Ground-truth mask; (c): Fist 6 latent feature maps; (d): Segmentation results.

Fig. 9 An example of segmentation for an image from COCO

As shown in (d), the person, pizza, table, and one of the cups are accurately segmented when compared with the ground-truth masks in (b). The results in (c) represent the first six latent feature maps obtained via SVD from the resized and concatenated feature maps of the stem and block 16 of EfficientNetB0. As observed in (c), the regions corresponding to the person, pizza, and one of the cups are significantly attended.

Fig. 10 shows an example of segmentation for an image from ADE2k, with an *HIoU* of 64.6%.

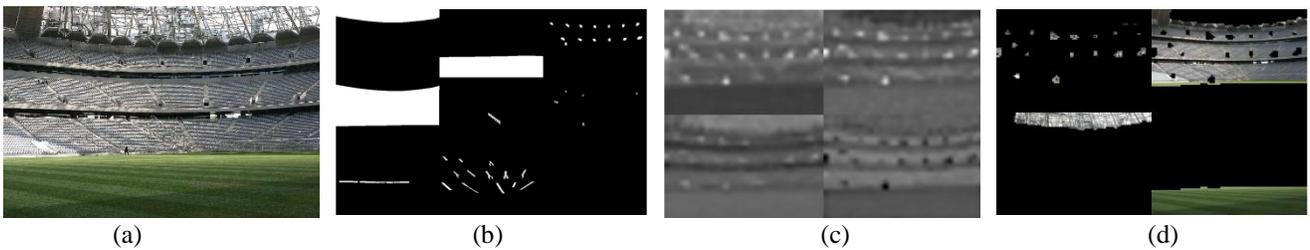

(a): Original image; (b): Ground-truth mask; (c): Fist 4 latent feature maps; (d): Segmentation results.

Fig. 10 An example of segmentation for an image from ADE2K

As shown in (d), besides the grassland, grandstand, and eaves, small objects such as persons and entrances are also accurately segmented, while the corresponding regions are significantly activated according to the latent feature maps shown in (c).

Fig. 11 shows an example of segmentation for an image from PASCAL VOC, with an *HIoU* of 56.1%.

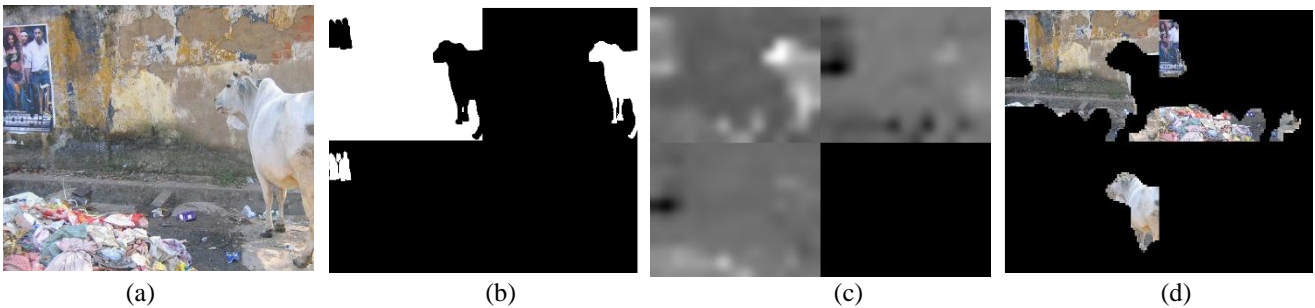

(a)                    (b)                    (c)                    (d)

(a): Original image; (b): Ground-truth mask; (c): Fist 3 latent feature maps; (d): Segmentation results.

Fig. 11 An example of segmentation for an image from PASCAL VOC

Compared with the ground truth masks shown in (b), the cow and persons are accurately segmented. The segmented regions remain precise and well-aligned with the ground truth mask, indicating strong segmentation performance even under cluttered outdoor conditions. However, objects such as the garbage dump, which are annotated as background in the ground truth masks, are also segmented together with the persons. As shown in (c), the region corresponding to the garbage dump is distinctly activated.

Table 1 summarizes the results of the state-of-the-art (SOTA) methods and our proposed approaches on the COCO-Obj-80, ADE20K, and PASCAL VOC benchmarks. The values in the table represent the mean $HIoU$ scores computed over all images in each dataset. OVSR1 and OVSR2 denote our proposed methods. OVSR1 corresponds to the case where the feature maps are not resized prior to singular value decomposition (SVD), whereas OVSR2 corresponds to the case where the feature maps are enlarged to a size of $64 \times (width/height) \times 64$ before SVD. As shown in Table 1, OVSR2 consistently achieves higher scores than OVSR1 across all three benchmarks, indicating that enlarging the feature maps enhances segmentation performance.

Furthermore, our methods outperform the SOTA approaches on COCO-Obj-80 and ADE20K, achieving mean $HIoU$ scores of 35.3 % and 37.2 % on COCO-Obj-80, and 44.5 % and 45.1 % on ADE20K. For COCO-Obj-80, the improvement over the third-highest SOTA score is 3.1 %, whereas for ADE20K, it is 0.8 %. However, our methods rank third and fourth on the PASCAL VOC benchmark, achieving mean $HIoU$ scores of 57.7% and 58.7%, respectively. We infer that the lower scores compared with the SOTA methods arise because certain regions annotated as background in VOC are activated by the models and segmented as objects. Fig. 11 provides a representative example of this phenomenon. Next, we plan to investigate why the DiffCut and YONG models, particularly YONG, achieve significantly higher $HIoU$ scores on VOC, yet lower scores than ours on COCO-Obj-80 and ADE20K.

Table 1 Mean $HIoU$ of SOTA methods and ours

| Model | VOC | COCO-Obj-80 | ADE20K |
|---|---|---|---|
| ReCO [30] | 25.1 | 15.7 | 11.2 |
| MaskCLIP [31] | 38.8 | 20.6 | 9.8 |
| MaskCut [32] | 53.8 | 30.1 | 35.7 |
| DiffSeg [33] | 48.2 | 31.7 | 39.9 |
| DiffCut [13] | 65.2 | 34.1 | 44.3 |
| PGCVLM [15] | 52.2 | 32 | 13.5 |
| YONG [34] | 94.7 | for training | 31.1 |
| OVSR1(ours) | 57.7 | 35.3 | 44.5 |
| OVSR2(ours) | 58.7 | 37.2 | 45.1 |

*C. Object recognition results*

Open-vocabulary recognition experiments are also conducted on the COCO-Obj-80, PASCAL VOC, and ADE20K datasets.

First, the impact of three types of phrases on object recognition is investigated using COCO-Obj-80. The performance in terms of accuracy, precision, recall, F1-score, and average precision (AP) for the three prompt types (Phrase 1–3) under both the original and SVD conditions is summarized in Tables 2–4. The category names follow those defined in COCO, with an additional category labeled "something else."

Table 2 Results of Phrase 1

| Prompt | SVD | Accuracy | Recall | Precision | F1 | AP |
|---|---|---|---|---|---|---|
| Phrase1 | X | 0.429 | 0.601 | 0.587 | 0.544 | 35.2 |
| Phrase1 | O | 0.32 | 0.484 | 0.671 | 0.499 | 32.5 |

Table 3 Results of Phrase 2

| Prompt | SVD | Accuracy | Recall | Precision | F1 | AP |
|---|---|---|---|---|---|---|
| Phrase2 | X | 0.396 | 0.577 | 0.637 | 0.562 | 36.4 |
| Phrase2 | O | 0.448 | 0.533 | 0.619 | 0.505 | 33 |

Table 4 Results of Phrase 3

| Prompt | SVD | Accuracy | Recall | Precision | F1 | AP |
|---|---|---|---|---|---|---|
| Phrase3 | X | 0.386 | 0.545 | 0.744 | 0.588 | 40.5 |
| Phrase3 | O | 0.493 | 0.598 | 0.465 | 0.461 | 27.8 |

As presented in Table 2–4, the model using Phrase3 without SVD achieved the best overall performance, with the highest F1-score and AP. Among the three prompt templates, the reason that Phrase3 which is the simplest form without a super-category yields the best overall performance is as its direct phrasing prevented semantic interference from broader contextual terms.

Phrase2 achieved balanced results by combining descriptive structure with categorical information, while Phrase1, which heavily relied on super-categories, exhibited reduced fine-grained recognition accuracy, particularly after SVD. This indicates that super-category information has dual effects: beneficial for broad generalization but detrimental to distinguishing fine-grained categories.

Based on these findings, Phrase3 is adopted as the standard prompt setting for subsequent experiments on PASCAL VOC and ADE20K, as well as for SOTA comparisons.

Table 5 presents the results of the proposed method with Phrase 3 on VOC.

Table 5 Results on VOC

| Prompt | SVD | Accuracy | Recall | Precision | F1 | AP |
|--------|-----|----------|--------|-----------|-----|-----|
| Phrase3 | X | 0.683 | 0.78 | 0.89 | 0.802 | 69.4 |
| Phrase3 | O | 0.754 | 0.83 | 0.513 | 0.606 | 42.5 |

Table 6 presents the results of the proposed method with Phrase 3 on ADE20K.

Table 6 Results on ADE20K

| Prompt | SVD | Accuracy | Recall | Precision | F1 | AP |
|--------|-----|----------|--------|-----------|-----|-----|
| Phrase3 | X | 0.25 | 0.365 | 0.329 | 0.267 | 12 |
| Phrase3 | O | 0.261 | 0.373 | 0.304 | 0.23 | 11.3 |

For VOC, which contains the fewest 20 category labels, all performance metrics reach their highest values. In contrast, for ADE20K, which includes the most 150 categories, all performance metrics attain their lowest values. This observation highlights that CLIP's zero-shot generalization strongly depends on the semantic separability of text embeddings. Datasets with fewer, well-defined categories (like VOC) align better with CLIP's training distribution, whereas dense or fine-grained taxonomies (like ADE20K) exceed the discriminative granularity that CLIP's text embeddings can reliably support.

Moreover, as shown in Tables 4–6, the accuracy and recall increase after applying SVD, whereas the precision decreases across all three datasets. For VOC, both the increase and decrease are substantial, whereas for ADE20K, the changes are within approximately 2%. VOC shows a larger variation because it contains fewer and more distinct categories, making CLIP's text embeddings highly separable. Applying SVD amplifies dominant semantic features, enhancing alignment and recall but reducing precision through overactivation. ADE20K, with many fine-grained, overlapping categories, exhibits weaker effects, keeping performance changes within about 2%. Generally, this occurs because SVD preserves dominant semantic components in the visual features, enhancing alignment with CLIP text embeddings, which improves recall and overall accuracy. However, the suppression of low-variance components reduces fine-grained discriminability, causing overlapping activations among semantically similar categories and thus lowering precision. In essence, SVD functions, as a semantic amplifier, it strengthens global object–category correspondence but weakens category specificity. When applied appropriately based on dataset characteristics, it effectively enhances accuracy and generalization in open-vocabulary recognition without additional training.

To further assess effectiveness, the proposed method was compared with SOTA approaches, including ViLD [16], MarvelOVD [20], HD-OVD [18], DK-DETR [17], and MaskCLIP [35]. Table 7 summarizes the mAP results of SOTA research and ours on three benchmarks.

Table 7 Comparison with SOTA approaches

| Model | COCO-obj-80 | VOC | ADE20K |
|-------|-------------|-----|--------|
| **Extra training** | | | |
| ViLD[16] | 36.6 | 72.2 | —— |
| MarvelOVD[20] | 38.9 | —— | —— |
| HD-OVD[18] | 36.6 | —— | —— |
| DK-DETR[17] | 39.4 | 71.3 | —— |
| MaskCLIP[35] | —— | —— | 6.164 |
| **training-free** | | | |
| OVSR(ours) | 40.5 | 69.4 | 12 |

Our method, OVSR, without applying SVD, outperforms the SOTA approaches on COCO-Obj-80 and ADE20K, achieving *mAP* scores of 40.5% and 12.0%, respectively. For COCO-Obj-80, the improvement over the second-highest SOTA score is 1.1%, whereas for ADE20K, it is 5.8%. However, on the PASCAL VOC benchmark, our method attains a *mAP* of 69.4%, which is lower than the SOTA results of 72.2% and 71.3%, respectively. We infer that this occurs because the performance of object segmentation on VOC is lower than that of the SOTA methods.

Moreover, it should be emphasized that the proposed framework is training-free, whereas the SOTA approaches require additional training procedures, such as knowledge distillation, re-weighted learning, and semantic transfer. This demonstrates the flexibility, generalizability, and practicality of the proposed method.

Some results of recognizing objects in images from COCO-Obj-80, PASCAL VOC, and ADE20K are shown in Fig. 12.

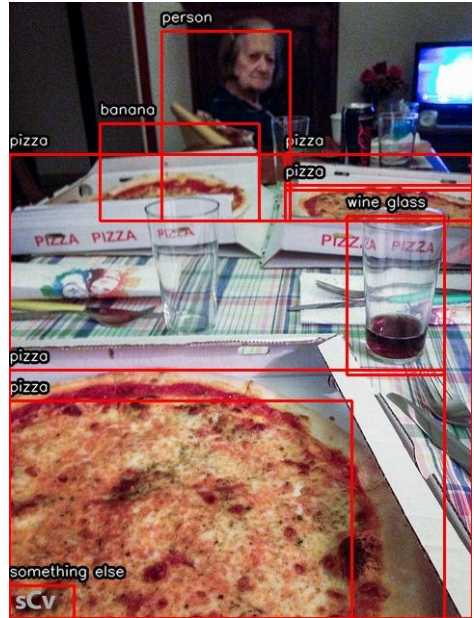

(a) from COCO

ground truth: cup, dining table, fork, knife, person, pizza, tv

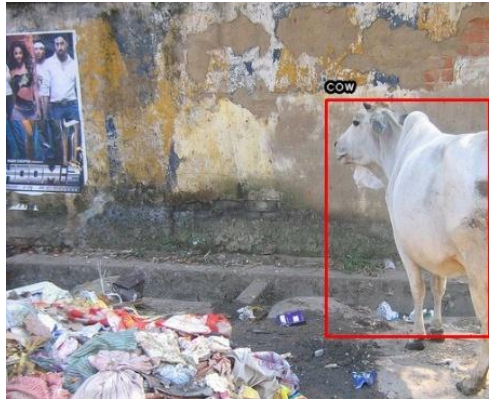

(b) VOC

ground truth: cow, person

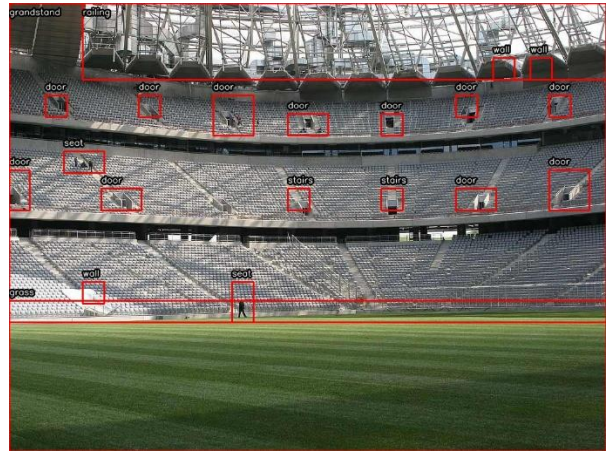

(C)ADE20K

ground truth: banister, ceiling, field, gate, grandstand, covered stand, person, railing, stairs

Fig. 12 Visualization results

For the COCO example, the ground truth labels include categories such as cup, dining table, fork, knife, person, pizza, and tv. The proposed method successfully identifies person and pizza as primary instances, while also detecting related contextual objects such as wine glass and banana, which are semantically associated but not explicitly labeled in the ground truth. Moreover, the lower-left region with characters is identified as something else. This suggests that the model can generalize to semantically relevant but unlabeled objects, reflecting its open-vocabulary property.

For the VOC example, the ground truth labels contain cow and person. The model correctly identifies the cow but misses the person.

For the ADE20K example, the ground truth labels include banister, ceiling, field, gate, grandstand, covered stand, person, railing, and stairs. The proposed model successfully identifies railing, grandstand, and stairs. Furthermore, seat and door, which semantically correspond to the ground-truth categories of covered stand and gate, respectively, are also identified. This result demonstrates that the model tends to capture semantically similar concepts even when exact label names differ. Such behavior reflects the strength of CLIP-based text embeddings in understanding cross-category semantic relationships.

However, it should be noted that certain limitations remain in object and category encoding when directly utilizing CLIP. Next, we plan to conduct contrastive learning by combining pixel-wise feature representations derived from convolutional neural networks with text embeddings obtained from large language models to enhance the performance of open-vocabulary object recognition.

## VI. Conclusion

In this work, we presented a training-free framework for open-vocabulary image segmentation and recognition that integrates convolutional feature extraction, SVD, and vision–language modeling. By constructing pixel-level representations from multi-scale EfficientNetB0 features and decomposing them into a compact latent space, our method enables unsupervised segmentation through hierarchical clustering with automatically determined cluster numbers. Furthermore, by leveraging CLIP-based text–image embeddings, we achieve open-vocabulary object recognition without additional training or fine-tuning. Experimental results on multiple benchmarks demonstrate that our approach not only outperforms existing state-of-the-art methods on COCO-Obj-80 and ADE20K but also provides competitive performance on PASCAL VOC. The results confirm the flexibility, generalizability, and practicality of our training-free framework. Future work will focus on enhancing category discriminability through contrastive learning that combines pixel-wise visual features with text embeddings derived from large language models.

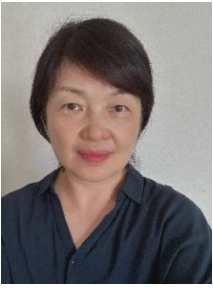

**Ying Dai** (M'98) Dr. Ying Dai received her B. S. and M.S. degrees from Xian Jiaotong University, China in 1985 and 1988, respectively. After some years working in the same university, she attended Department of Information Engineering, Shinshu University, Japan in 1992. She had a Dr. Eng degree from Shinshu University in 1996. She was granted JSPS Research Fellowships for Young Scientists from 1995 to 1997. She joined Iwate Pref. University in 1998. She has been a professor in the Faculty of Software and Information Science, Iwate Pref. University. Her main research interests are in the area of pattern recognition, image understanding, Kansei information processing, and computational intelligence.

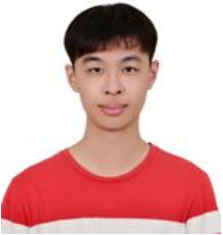

**Wei Yu Chen** Wei Yu Chen received his master's degree in Computer Science and Information Engineering from Chaoyang University of Technology, Taiwan in 2024. He is currently pursuing a double master's degree at the Graduate School of Software and Information Science, Iwate Prefectural University, Japan. His research interests include image processing, object recognition, and the Internet of Things (IoT). His current work focuses on open-vocabulary object recognition and semantic understanding based on vision-language models.